\documentclass[sigconf]{acmart}

\copyrightyear{2026}
\acmYear{2026}
\setcopyright{cc}
\setcctype{by}
\acmConference[WWW '26] {Proceedings of the ACM Web Conference 2026}{April 13--17, 2026}{Dubai, United Arab Emirates.}
\acmBooktitle{Proceedings of the ACM Web Conference 2026 (WWW '26), April 13--17, 2026, Dubai, United Arab Emirates}
\acmISBN{979-8-4007-2307-0/2026/04}
\acmDOI{10.1145/XXXXXX.XXXXXX}
\usepackage{graphicx}
\usepackage{subcaption}
\usepackage{amsmath, amsfonts}
\usepackage{algorithm}
\usepackage{algpseudocode}
\usepackage{multirow}
\usepackage{array}
\usepackage{xcolor}
\usepackage{float}

\usepackage{enumitem}
\usepackage{hyperref}

\usepackage[font=small,skip=0pt]{caption}
\makeatletter
\def\@secfont{\bfseries\Large\section@raggedright}
\def\@subsecfont{\bfseries\section@raggedright}
\def\@subsubsecfont{\itshape\section@raggedright}
\makeatother

\title{\vspace{-1mm}Machine Learning as a Service (MLaaS) Dataset Generator Framework for IoT Environments}

\author{Deepak Kanneganti}
\email{s.kanneganti@curtin.edu.au}
\affiliation{%
  \institution{Curtin University}
  \city{Perth}
  \country{Australia}
}
\author{Sajib Mistry}
\email{sajib.mistry@curtin.edu.au}
\affiliation{%
  \institution{Curtin University}
  \city{Perth}
  \country{Australia}
}
\author{Sheik Fattah}
\email{sheik.fattah@curtin.edu.au}
\affiliation{%
  \institution{Curtin University}
  \city{Perth}
  \country{Australia}
}
\author{Joshua Boland}
\email{joshua.boland@student.curtin.edu.au}
\affiliation{%
  \institution{Curtin University}
  \city{Perth}
  \country{Australia}
}
\author{Aneesh Krishna}
\email{aneesh.krishna@curtin.edu.au}
\affiliation{%
  \institution{Curtin University}
  \city{Perth}
  \country{Australia}
}

\begin{document}

\begin{abstract}
We propose a novel MLaaS Dataset Generator (MDG) framework that creates configurable and reproducible datasets for evaluating Machine Learning as a Service (MLaaS) selection and composition. MDG simulates realistic MLaaS behaviour by training and evaluating diverse model families across multiple real-world datasets and data distribution settings. It records detailed functional attributes, quality of service metrics, and composition-specific indicators, enabling systematic analysis of service performance and cross-service behaviour. Using MDG, we generate more than ten thousand MLaaS service instances and construct a large-scale benchmark dataset suitable for downstream evaluation. We also implement a built-in composition mechanism that models how services interact under varied Internet of Things conditions. Experiments demonstrate that datasets generated by MDG enhance selection accuracy and composition quality compared to existing baselines. MDG provides a practical and extensible foundation for advancing data-driven research on MLaaS selection and composition.

\end{abstract}




\maketitle

\section{Introduction}


\newcounter{sagemakerfn}

Machine Learning as a Service (MLaaS) has become a central component of modern cloud and Internet of Things ecosystems. It offers scalable tools to train, deploy, and manage machine learning models ~\cite{ribeiro2015mlaas}. Major providers such as \textit{Microsoft Azure}\footnote{\url{https://azure.microsoft.com/en-us/products/machine-learning/}}, \textit{AWSSageMaker}\footnote{\url{https://aws.amazon.com/sagemaker/}}\setcounter{sagemakerfn}{\value{footnote}}, and \textit{OpenAI ChatGPT API}\footnote{\url{https://platform.openai.com/docs/}} offer comprehensive MLaaS solutions. These services enable organisations and devices to access advanced analytics without maintaining local infrastructure. MLaaS is becoming more useful in domains such as healthcare, transportation, and IoT applications. Effective selection and composition of MLaaS offerings remain difficult in practice despite such widespread adoption \cite{kanneganti2025adaptivecompositionmachinelearning}. Research in this area is highly dependent on high-quality datasets \cite{kanneganti2025adaptivecompositionmachinelearning, xie2022cost}. For example, evaluating a selection or composition algorithm requires detailed information on service accuracy, latency, reliability, and interactions between services. Reproducible benchmarks or fair comparisons between competing techniques cannot be developed without a high-quality dataset that captures these characteristics.

Existing MLaaS datasets are static, sparse, and not designed to capture the dynamic behaviour of modern cloud based machine learning services. Most available datasets contain only high level metadata, such as input type, accuracy, or pricing ~\cite{ribeiro2015mlaas}\cite{wang2024hsc}. They typically lack essential information about deeper functional attributes, resource usage, service reliability, and composability characteristics. Many of these datasets are outdated and numerous services are no longer available. None of these datasets provides insight into how different ML models interact when combined or reused in tasks. Collecting real MLaaS performance data is difficult due to provider restrictions, variable pricing, and the continuous evolution of service offerings. As a result, it is quite challenging to benchmark new selection or composition techniques, reproduce experimental conditions, and generate controlled scenarios that reflect realistic IoT demands. These limitations highlight the need for a configurable and reproducible way to generate MLaaS datasets rather than relying on fixed and incomplete data sources.

\begin{figure*}[t]
  \centering
  \includegraphics[width=\linewidth]{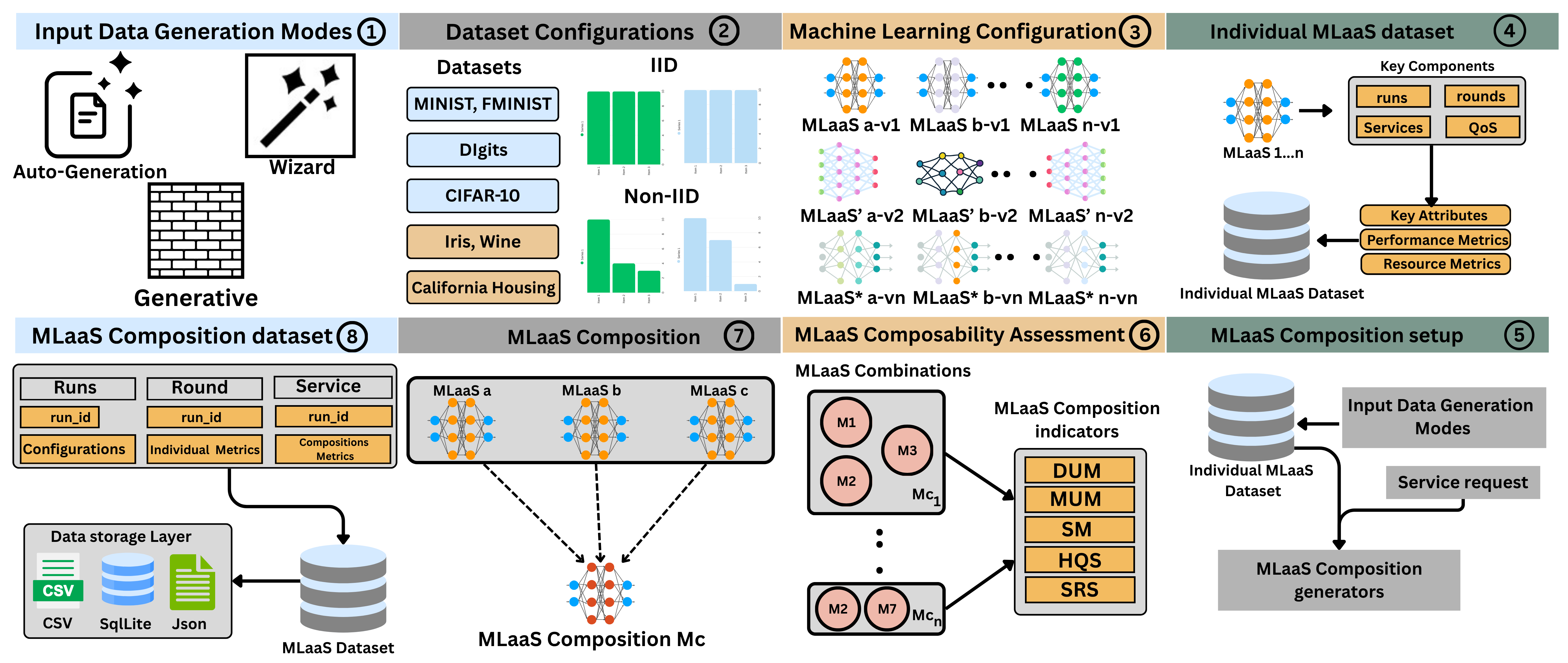}
  \caption{MLaaS Dataset Generator Framework}
  \label{fig1}
  \vspace{-5mm}
\end{figure*}

We present \textit{the MLaaS Dataset Generator (MDG)}, a framework designed to create configurable and reproducible datasets that capture realistic MLaaS behaviour for Internet of Things environments. Rather than relying on fixed or incomplete data sources, MDG simulates the operation of machine learning services by training and evaluating diverse models across multiple real-world datasets and configuration settings. It produces detailed functional attributes, quality of service metrics, and composition specific indicators that reflect how services perform both individually and when combined. MDG supports a wide range of model families, data distributions, and task types, enabling researchers to construct controlled scenarios that reflect heterogeneous and evolving IoT conditions. By automating dataset creation and incorporating service interaction modelling, the framework provides a practical and scalable foundation for evaluating MLaaS selection, predicting performance, and analysing composition strategies under consistent and reproducible conditions. In this work, we use the MDG framework to generate the most comprehensive MLaaS dataset to date and to enable systematic evaluation of service selection and composition techniques. MDG provides a flexible mechanism for constructing large and diverse sets of MLaaS service instances, capturing both individual performance and cross service behaviour under varied Internet of Things conditions. The framework also supports reproducible experimentation by exporting all generated information in structured formats suitable for downstream analysis. The main contributions of this paper are as follows:

\begin{itemize}[itemsep=0ex, leftmargin=2ex]
    \item We propose a novel MLaaS dataset generator (MDG) that reconstructs realistic service behaviour by capturing detailed functional attributes, QoS metrics, and composition-specific indicators across classification, regression, and clustering tasks.
    \item We develop a large-scale MLaaS dataset containing more than ten thousand service instances derived from multiple Internet of Things datasets, diverse model families, and varied data distribution settings.
    \item We introduce a built-in composition mechanism that evaluates how MLaaS services interact, enabling analysis of interdependencies, behavioural variation, and composability quality under controlled conditions.
    \item A comprehensive experimental study demonstrating that datasets produced by MDG improve service selection accuracy and composition quality compared with existing baselines.
\end{itemize}
\vspace{-3mm}
\section{MLaaS Dataset Generator Framework} The proposed \textit{MLaaS Dataset Generator (MDG)} framework reconstructs a cloud-scale environment for generating realistic MLaaS datasets tailored to IoT environments. Unlike static and incomplete datasets, MDG offers a configurable and reproducible mechanism for generating service instances supporting MLaaS selection and composition research. \textit{Figure~\ref{fig1}} illustrates the sequential stages of the end-to-end MLaaS dataset generation framework.

\vspace{-3mm}
\subsection{Input Data Generation Modes}
IoT environments require datasets that reflect changing application goals and service constraints. The framework offers three input data generation modes to support reproducible experiments and scalable benchmarks.

\begin{itemize}[itemsep=0ex, leftmargin=2ex]
    \item \textbf{Wizard Mode:} Designed for interactive and exploratory use. It guides users through dataset selection, model type, and training configuration via a prompt-driven interface, providing sensible defaults for quick setup and demonstration.
    \item \textbf{Generate Mode:} Enables reproducible experiments through command-line execution. Users can specify parameters such as dataset, model type, hyperparameters, making it suitable for controlled and repeatable batch runs.
    \item \textbf{Autogen Mode:} Automates large-scale input generation by randomly sampling configurations across model families, optimizers, and data-partition strategies. It supports benchmark creation and diverse scenario testing without manual intervention.
\end{itemize}

\subsection{Dataset and Machine Learning Configuration}
The MDG framework prepares datasets suitable for the selected task and target environment using the chosen input data generation mode. Table~\ref{tab1} lists the publicly available datasets supported by MDG, which are automatically normalised, preprocessed, and reshaped for \textit{classification}, \textit{regression}, or \textit{clustering} tasks. This automated preparation ensures consistency and reproducibility while still allowing flexible configuration when needed. To emulate realistic IoT conditions, MDG supports both \textit{Independent and Identically Distributed (IID)} and \textit{non-Independent and Identically Distributed (non-IID)} data strategies. As illustrated in Table~\ref{tab1}, data can be distributed among clients using Dirichlet, quantity-skew, or shard-based sampling, capturing natural imbalance and cross-client variability typical in IoT deployments. The framework supports multiple model architectures that are dynamically instantiated based on the chosen task and model family. As shown in Table~\ref{tab1}, available models include \textit{CNN, RNN, ANN, MLP, Logistic Regression, MobileNet, and Random Forest}, enabling the generation of diverse MLaaS service instances spanning different application domains and complexity levels. All execution details are recorded for downstream analysis: client-level information is stored in the \texttt{clients} table and run-level information in the \texttt{runs} table, ensuring a structured trace for selection, composition, and performance evaluation.

\begin{table}[t]
\centering
\caption{Datasets, ML Models, and Training Configuration Summary}\label{tab1}
\resizebox{0.40\textwidth}{!}{
\begin{tabular}{p{2.9cm} p{3.4cm}}
\toprule
\textbf{Datasets (7)} & \textbf{Model Families (6)} \\
\midrule
MNIST, Fashion-MNIST, Digits & CNN, RNN (LSTM/GRU), ANN/MLP, Logistic Regression \\
CIFAR-10 & MobileNetV2, CNN, MLP \\
Iris, Wine & MLP, Logistic Regression, Random Forest, K-means \\
California Housing & Random Forest, MLP, K-means \\\midrule
\textbf{Task Types (3)} & Classification, Regression, Clustering \\
\textbf{Service Instances} & 10,432 total services \\\midrule
\textbf{Data Distributions} & IID and non-IID (Dirichlet, shard, quantity-skew) \\
\textbf{Rounds per Run} & 5–50 rounds (avg. 20) \\
\textbf{Compositions} & 740 MLaaS service compositions \\
\bottomrule
\end{tabular}
}
\vspace{-3mm}
\end{table}

\subsection{Individual MLaaS dataset}

The \textit{Individual MLaaS Dataset} component of the MDG framework captures detailed metrics describing both model-level performance and system-level resource behaviour for each simulated MLaaS service. It enables controlled and reproducible evaluation of individual MLaaS services by logging an execution trace across multiple stages—\texttt{runs}, \texttt{rounds}, and \texttt{clients}—as shown in Figure~\ref{fig1}. The \texttt{runs} table stores configuration details such as dataset, task type, and model architecture, ensuring that a given service configuration can be regenerated consistently for repeated experimentation. The \texttt{rounds} and \texttt{clients} tables record global statistics (e.g., loss, participating clients, scheduling) and individual participation data (e.g., sample count and data distribution), reflecting the heterogeneous workload patterns common in IoT environments. QoS and behavioural metrics are maintained in the \texttt{qos} table, linking each service instance to its corresponding \texttt{run\_id}, \texttt{round}, and \texttt{client\_id}. These metrics include \textit{Performance Metrics} (accuracy, F1, precision, recall, RMSE, silhouette, inertia) and \textit{Resource Metrics} (computation time, CPU/GPU utilisation, memory usage, communication bytes). Performance metrics indicate predictive capability across supported task types, while resource metrics provide insight into efficiency, scalability, and operational cost under varied IoT workloads. Together, these measurements establish a comprehensive view of individual MLaaS behaviour, supporting systematic benchmarking and comparison of ML services under consistent and reproducible experimental conditions.


\subsection{MLaaS Composition setup}

The \textit{Individual MLaaS Dataset} generated in the previous stage serves as the foundation for constructing MLaaS composition scenarios. Each service instance contains functional and QoS attributes that help identify suitable combinations that reflect IoT application requirements. In this framework, user-specific conditions, such as accuracy, latency, and resource constraints, may be integrated into the setup to shortlist candidate services for composition. Existing studies on adaptive MLaaS composition provide benchmark strategies for orchestrating such service interactions~\cite{kanneganti2025adaptivecompositionmachinelearning, xie2024skyml}. These studies highlight the need for structured mechanisms that operate on pre-generated combinations rather than relying on direct manual composition.

\subsection{MLaaS Composability Assessment}
Manual evaluation of all possible combinations is computationally expensive , the composition process requires filtering out services that are unlikely to cooperate effectively. The dataset generation framework incorporates utility-driven indicators derived from prior studies~\cite{kanneganti2025adaptivecompositionmachinelearning, xie2024skyml, li2020federated}. These include:
\textit{Data Utility Measurement (DUM)}, \textit{Model Utility Measurement (MUM)}, \textit{Scalability Measurement (SM)}, \textit{Historical Quality Score (HQS)}, and \textit{Service Reliability Score (SRS)}. These metrics assist in identifying promising service configurations without composing every candidate pair or group, which can be computationally expensive in large-scale IoT environments. The goal of these indicators is to retain combinations with higher expected stability and performance while discarding unsuitable ones.

\subsection{MLaaS Composition}
Once composability filtering is complete, the framework performs MLaaS composition by integrating multiple services using benchmark composition mechanisms. Parametric services (e.g., neural models) are aggregated through weighted parameter averaging, while non-parametric services (e.g., Random Forest, clustering models) are integrated using ensemble-style statistical aggregation, similar to strategies adopted in federated learning~\cite{li2020federated,kanneganti2025adaptivecompositionmachinelearning,xie2022cost}. This simulation process reflects cross-service knowledge integration and enables evaluation of MLaaS behaviour under heterogeneous IoT workloads. The resulting metrics are analysed through score distributions, selection outcomes, and correlation views, offering interpretable insights into service suitability for IoT-focused MLaaS composition.

\vspace{-3mm}
\subsection{MLaaS Composition Dataset Collection and Storage} The MDG framework stores the outputs of MLaaS compositions as a structured MLaaS composition dataset. It records global metrics (accuracy, latency, communication cost, computation time, resource usage) and composition statistics (client participation, model updates, composability indicators) in linked \texttt{runs}, \texttt{rounds}, \texttt{clients}, and \texttt{qos} tables within a \texttt{SQLite} backend. The dataset is exported in \texttt{SQLite}, \texttt{CSV}, and \texttt{JSON} formats.

\vspace{-3mm}
\section{Experiments and results}
In this section, we evaluate the average solution quality of benchmark MLaaS composition techniques using the dataset created by the proposed MDG framework. We first assess MLaaS service selection performance using three datasets: the existing QWS~\cite{al2008investigating}, the incomplete MLaaS dataset~\cite{patel2024context}, and our proposed dataset. Standard benchmark techniques, including Rule-based~\cite{rhayem2021semantic}, Distance-based~\cite{patel2023machine}, and Skyline-based~\cite{patel2024context} techniques, are applied to evaluate overall selection quality. We then measure the average solution quality of MLaaS compositions generated using the benchmark composability model to demonstrate that the proposed dataset enhances both composition quality and performance across MNIST, Fashion-MNIST, and Digits datasets. All experiments were conducted on an Intel Core i7 machine (16~GB RAM) using Python, with results and source code available in our repository\footnote{https://github.com/YGYerrd/Adaptive-MLaaS-Model/tree/file-seperation}.



\subsection{Experiment 1: Evaluating MLaaS Service Selection Performance Across Benchmark Datasets}

\begin{table}[t]
\centering
\caption{Evaluating the performance of MLaaS service selection techniques using the dataset generated by the proposed MDG framework and benchmark datasets.}
\label{tab:req_satisfaction}
\begin{tabular}{lccc}
\toprule
\textbf{Technique} & \textbf{QWS\cite{al2008investigating}} & \textbf{In-MLaaS\cite{kanneganti2025adaptivecompositionmachinelearning}} & \textbf{Our Dataset}\\
\midrule
Rule-based \cite{rhayem2021semantic}   &0.82 & 0.85 & \textbf{0.92} \\
Distance-based \cite{patel2023machine}  &0.88 &  0.96 &\textbf{0.99} \\
Skyline-based \cite{patel2024context}  &0.50 &  0.70 &\textbf{0.81} \\
\bottomrule
\end{tabular}
\end{table}

The MDG framework support the researchers with MLaaS dataset useful for MLaaS service selection and composition research problems for IoT environments. To evaluate the effectiveness of the dataset generated by the MDG framework in supporting MLaaS service selection, we compare its performance using traditional selection techniques such as rule-based methods \cite{rhayem2021semantic}, distance-driven selection approaches \cite{patel2023machine}, and skyline-oriented service filtering \cite{patel2024context}. Table~\ref{tab:req_satisfaction} compares the results obtained using our MLaaS dataset with the existing benchmark datasets, including QWS~\cite{al2008investigating} and the incomplete MLaaS dataset~\cite{patel2024context}. In this experiment, service requests were formed to reflect typical IoT needs, including targets functional and QoS attributes. The results shows that the rule-based, distance-based, and skyline-based methods achieve satisfaction rates of 0.92, 0.99, and 0.81, respectively, when using the proposed dataset, demonstrating a clear advantage in accuracy and reliability of service selection for IoT applications. In addition, the incomplete MLaaS dataset provides only limited QoS visibility, making it difficult to evaluate service behaviour under realistic IoT constraints and resulting in less reliable selection outcomes.  
Experiments results highlights that the proposed dataset delivers an average performance gain of \textbf{15–25\% }over existing datasets, showing its accuracy in solving MLaaS service selection problem.

\subsection{Experiment 2: Evaluating Average Solution Quality of MLaaS Compositions Using the Proposed Dataset}
\begin{figure}[t]
    \centering

    \begin{subfigure}[b]{0.90\columnwidth}
        \centering
        \includegraphics[width=\linewidth]{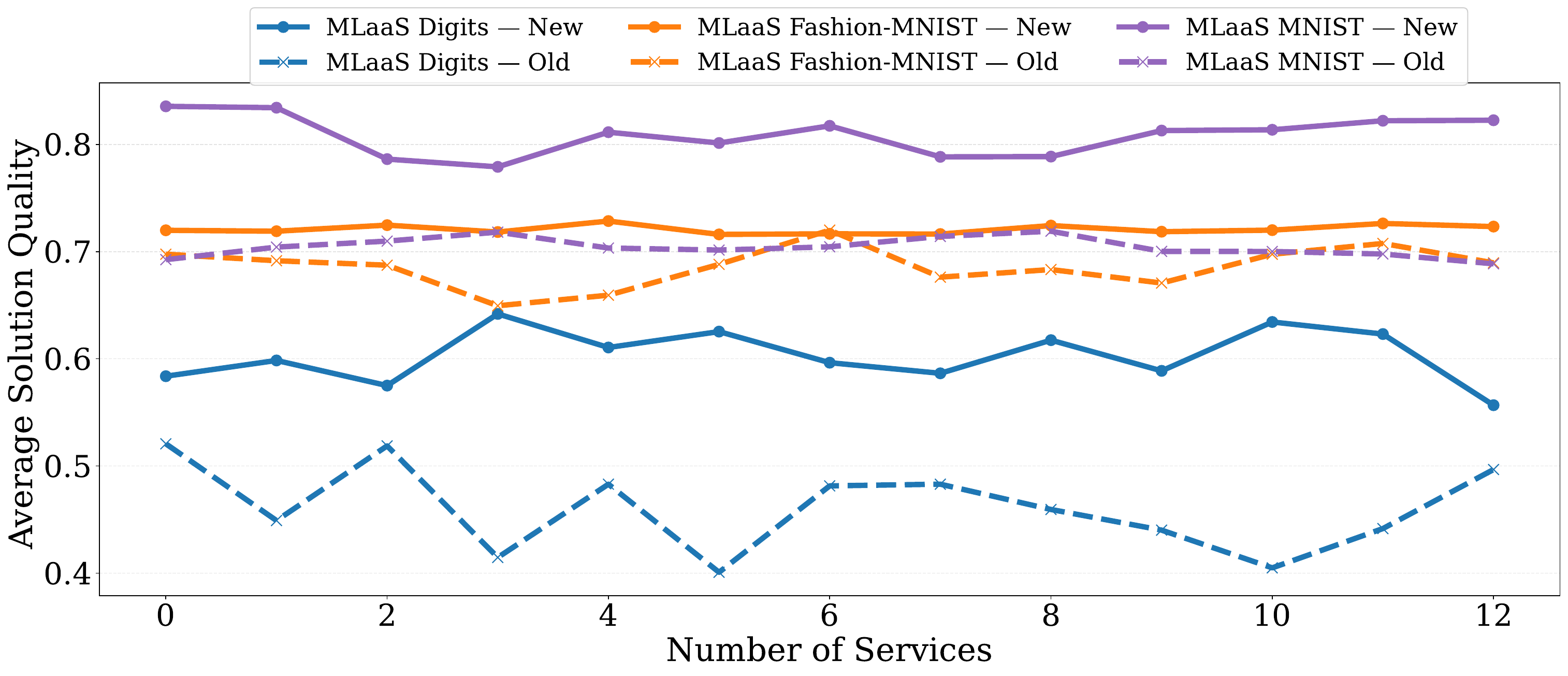}
        \caption{}
        \label{fig:mlaas_quality_a}
    \end{subfigure}

    \vspace{3mm}

    \begin{subfigure}[b]{0.90\columnwidth}
        \centering
        \includegraphics[width=\linewidth]{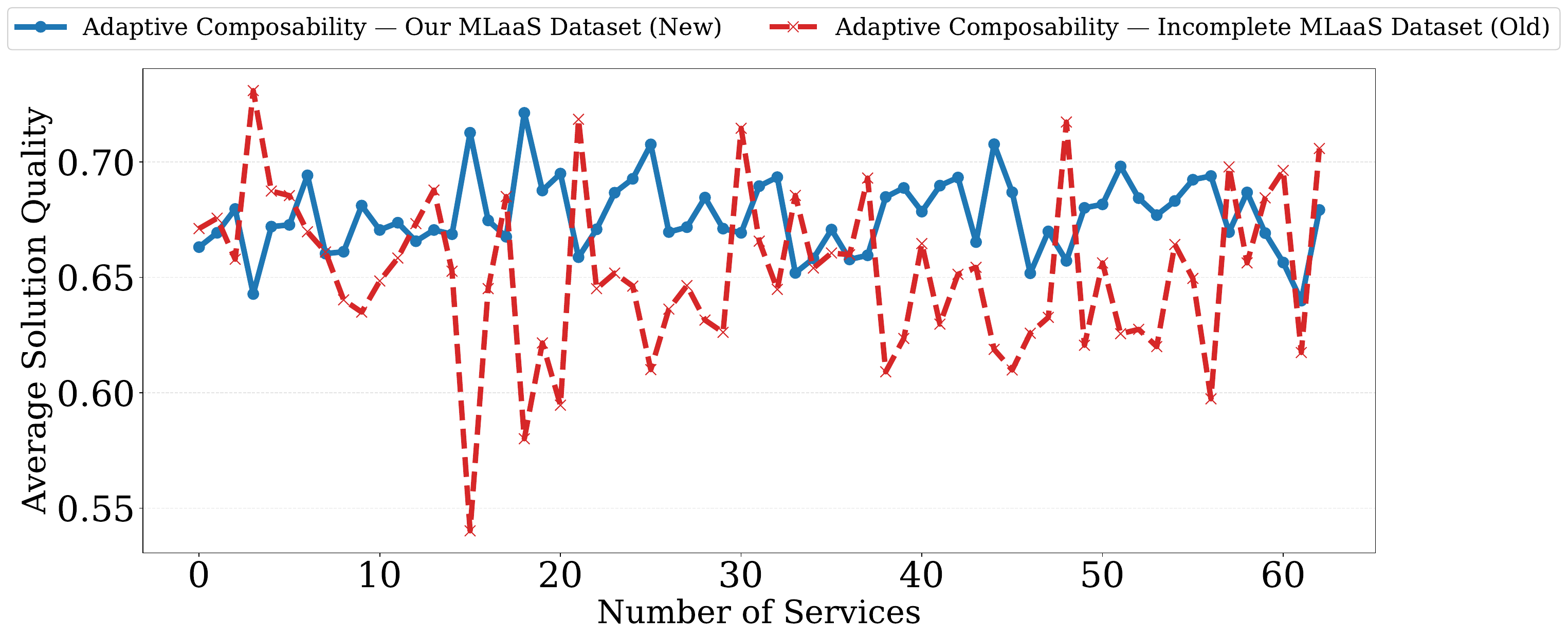}
        \caption{}
        \label{fig:mlaas_quality_b}
    \end{subfigure}

    \caption{Evaluation of MLaaS composition problem using the proposed MDG framework dataset: (a) Average solution quality across benchmark MLaaS datasets, (b) MLaaS composability performance comparison between the proposed and incomplete MLaaS datasets.}
    \label{fig:mlaas_quality}
    \vspace{-3mm}
\end{figure}

The MDG framework generates composition data from service combinations, which is used solely to assess composition behaviour and compare with existing datasets. Effective composition requires datasets that capture diverse characteristics of MLaaS services when integrated. To evaluate efficiency and average solution quality, we used benchmark MLaaS composition techniques\cite{kanneganti2025adaptivecompositionmachinelearning} and compared composability scores obtained from the existing incomplete dataset and our proposed dataset. Figure \ref{fig:mlaas_quality_a} shows the average solution quality for compositions using both datasets, where the x-axis represents the number of services and the y-axis denotes the composability score. Higher scores indicate stronger compatibility and stability between composed services. Due to space constraints, we limited experiments to MLaaS services derived from MNIST, Fashion-MNIST, and Digits. Figure \ref{fig:mlaas_quality_b} highlights that the traditional dataset exhibits fluctuating solution quality, often below 0.60, while the proposed dataset maintains a steadier trend, averaging around 0.68 compared to approximately 0.58. This stability indicates better adaptive composability with reduced performance fluctuations. Overall, the proposed dataset achieves an \textit{average quality solution gain of 10\%}, demonstrating its effectiveness in enabling realistic, high-quality MLaaS service selection and composition for diverse IoT requirements.

\section{Conclusion and Future Work}
We propose a novel MDG framework that produces benchmark datasets supporting both MLaaS service selection and composition, enabling diverse data generation modes that reflect real-world MLaaS configurations and composability scenarios. Experiments across MNIST, Fashion-MNIST, and Digits show 15–25\% improvement in service selection and about 10\% improvement in composition quality, demonstrating the framework’s value for data-driven MLaaS research. In future work, we plan to extend the dataset with a broader range of models and composability attributes, increasing its applicability to complex IoT environments.





\bibliographystyle{unsrt}
\bibliography{references}

@inproceedings{ribeiro2015mlaas,
  title={Mlaas: Machine learning as a service},
  author={Ribeiro, Mauro and et al.},
  booktitle={2015 IEEE 14th ICMLA},
  pages={896--902},
  year={2015},
  organization={IEEE}
}

@inproceedings{wang2024hsc,
  title={HSC: An artificial intelligence service composition dataset from hugging face},
  author={Wang, Xiao and et al. },
  booktitle={ICSOC},
  year={2024},
  organization={Springer}
}

@inproceedings{al2008investigating,
  title={Investigating web services on the world wide web},
  author={Al-Masri, Eyhab and Mahmoud, Qusay H},
  booktitle={World Wide Web},
  pages={795--804},
  year={2008}
}

@article{rhayem2021semantic,
  title={A semantic-enabled and context-aware monitoring system for the internet of medical things},
  author={Rhayem, Ahlem and et al.},
  journal={Expert Systems},
  year={2021}
}

@inproceedings{xie2022cost,
  title={Cost effective MLaaS federation: A combinatorial reinforcement learning approach},
  author={Xie, Shuzhao and Xue, Yuan and Zhu, Yifei and Wang, Zhi},
  booktitle={IEEE INFOCOM 2022},
  pages={2078--2087},
  year={2022},
  organization={IEEE}
}

@inproceedings{patel2024context,
  title={Context-Aware Selection of Machine Learning as a Service (MLaaS) in IoT Environments},
  author={Keya Patel and et al},
  booktitle={WISE},
  year={2024},
  organization={Springer}
}

@misc{kanneganti2025adaptivecompositionmachinelearning,
      title={Adaptive Composition of Machine Learning as a Service (MLaaS) for IoT Environments}, 
      author={Deepak Kanneganti and et al},
      year={2025},
      eprint={2506.11054},
      archivePrefix={arXiv},
      primaryClass={cs.LG},
      url={https://arxiv.org/abs/2506.11054}, 
}

@article{patel2023machine,
  title={Machine Learning as a Service (MLaaS) Selection with Incomplete QoS Information},
  author={Patel, Keya and et al.},
  year={2023}
}

@article{xie2024skyml,
  title={SkyML: A MLaaS Federation Design for Multicloud-based Multimedia Analytics},
  author={Xie, Shuzhao and et al.},
  journal={IEEE Transactions on Multimedia},
  year={2024},
  publisher={IEEE}
}

@article{li2020federated,
  title={Federated optimization in heterogeneous networks},
  author={Li, Tian and et al.},
  journal={Proceedings of Machine learning and systems},
  year={2020}
}

\end{document}